\documentclass[twocolumn, switch]{article} 

\usepackage{preprint}

\usepackage{amsmath, amsthm, amssymb, amsfonts}

\usepackage[numbers,square]{natbib}
\usepackage{natbib}

\usepackage[utf8]{inputenc}	
\usepackage[T1]{fontenc}	
\usepackage{xcolor}		
\usepackage[colorlinks = true,
            linkcolor = purple,
            urlcolor  = blue,
            citecolor = cyan,
            anchorcolor = black]{hyperref}	
\usepackage{booktabs} 		
\usepackage{nicefrac}		
\usepackage{microtype}		
\usepackage{lineno}		
\usepackage{float}			
\usepackage{amsmath}
\usepackage{multirow}
\usepackage{lipsum}		

\usepackage{enumitem}
\usepackage[table]{xcolor}
\usepackage{colortbl}        
\usepackage{xfp}

\usepackage[utf8]{inputenc}
\usepackage[T1]{fontenc}
\usepackage{tcolorbox}
\usepackage{listings}

\usepackage{algorithm}
\usepackage{algpseudocode}  

\usepackage[export]{adjustbox} 
\usepackage{subcaption}

\tcbuselibrary{listings, breakable}

\newtcblisting{promptbox}{
    arc=2pt,
    boxrule=0.5pt,
    colback=gray!5,
    colframe=black!80,
    left=4pt,
    right=4pt,
    top=0pt,
    bottom=0pt,
    boxsep=0pt,
    listing only,
    listing engine=listings,
    listing options={
        basicstyle=\ttfamily\small,
        breaklines=true,
        breakatwhitespace=false,
        keepspaces=true,
        columns=fullflexible,
        showstringspaces=false,
        breakindent=0pt,
        breakautoindent=false,
        xleftmargin=4pt,           
        xrightmargin=4pt,          
        resetmargins=true,         
    },
    breakable,
}

\usepackage[table]{xcolor}

\definecolor{myTextGreenColor}{HTML}{2CA02C}
\definecolor{myTextBenchColor}{HTML}{1F77B4}

\definecolor{myGreenColor}{HTML}{6ACC64}
\definecolor{myRedColor}{HTML}{D65F5F}

\newcommand{\ratioCell}[1]{%
    \ifdim #1pt > 0pt
      \cellcolor{myGreenColor!\fpeval{ min(100, max(0, (abs(#1) / 70) * 100)) }} #1
    \else\ifdim #1pt < 0pt
      \cellcolor{myRedColor!\fpeval{ min(100, max(0, (abs(#1) / 70) * 100)) }} #1
    \else
      #1
    \fi\fi
}

\usepackage{newfloat}
\DeclareFloatingEnvironment[name={Supplementary Figure}]{suppfigure}
\usepackage{sidecap}
\sidecaptionvpos{figure}{c}

\usepackage{titlesec}
\titlespacing\section{0pt}{12pt plus 3pt minus 3pt}{1pt plus 1pt minus 1pt}
\titlespacing\subsection{0pt}{10pt plus 3pt minus 3pt}{1pt plus 1pt minus 1pt}
\titlespacing\subsubsection{0pt}{8pt plus 3pt minus 3pt}{1pt plus 1pt minus 1pt}

\usepackage{tikz,xcolor,hyperref}

\definecolor{lime}{HTML}{A6CE39}
\DeclareRobustCommand{\orcidicon}{
	\begin{tikzpicture}
	\draw[lime, fill=lime] (0,0)
	circle [radius=0.16]
	node[white] {{\fontfamily{qag}\selectfont \tiny ID}};
	\draw[white, fill=white] (-0.0625,0.095)
	circle [radius=0.007];
	\end{tikzpicture}
	\hspace{-2mm}
}
\foreach \x in {A, ..., Z}{\expandafter\xdef\csname orcid\x\endcsname{\noexpand\href{https://orcid.org/\csname orcidauthor\x\endcsname}
			{\noexpand\orcidicon}}
}

\title{Dual Tuning for Reasoning Efficacy-Driven Data Curation in Multimodal LLM Training}

\author{
 Ruobing Zheng\textsuperscript{*},\\
  Ant Group\\
  \texttt{zhengruobing.zrb@antgroup.com} \\
  \And
 Tianqi Li\textsuperscript{*},\\
  Ant Group\\
  \texttt{shijian.ltq@antgroup.com} \\
  \And
 Jianing Li \\
  Ant Group\\
  \texttt{tiansuo.ljn@antgroup.com} \\
  \AND
 Qingpei Guo \\
  Ant Group \\
  \texttt{qingpei.gqp@antgroup.com} \\
  \And
 Yi Yuan \\
  Ant Group \\
  \texttt{yy464236@antgroup.com} \\
  \And
 Jingdong Chen \\
  Ant Group \\
  \texttt{jingdongchen.cjd@antgroup.com} \\
}

\begin{document}

\twocolumn[ 
  \begin{@twocolumnfalse} 

\maketitle

\centering{\textbf{Project Page:} \href{https://digital-avatar.github.io/ai/ThinkingBoundary/}{\textcolor{red}{https://digital-avatar.github.io/ai/ThinkingBoundary/}}}

\hspace{\fill}

\begin{abstract}
    Reasoning post-training improves Large Language Models (LLMs) on complex tasks such as mathematics and coding, but its benefits across diverse multimodal tasks remains uncertain. The trend of releasing parallel ``Instruct'' and ``Thinking'' models by leading teams is both resource-intensive and user-unfriendly. Prior work finds that the gains from reasoning training are influenced by multiple factors, such as base model capabilities, task characteristics, and Chain-of-Thought (CoT) data quality. However, principled criteria for determining when reasoning post-training is beneficial and which data should support it are still lacking. In this paper, we propose Dual Tuning, a reasoning efficacy-driven data curation framework for multimodal LLMs training. Given a target task and a base model, Dual Tuning jointly evaluates whether the training data is beneficial and whether reasoning training with current CoT content yields positive gains over non-reasoning alternatives. We apply Dual Tuning across spatial, mathematical, and multi-disciplinary tasks, and further analyze how reinforcement learning and thinking patterns affect reasoning efficacy. The Dual Tuning results guide data curation by identifying data that benefit reasoning training, data better suited to direct-answer training, and data that are detrimental under both training modes. Our work provides quantitative criteria for selecting appropriate training data and matching post-training strategies.
\end{abstract}
\vspace{0.35cm}

  \end{@twocolumnfalse} 
] 


\let\thefootnote\relax\footnotetext{*~Equal Contribution}


  

\section{Introduction}

Reasoning-enhanced large language models (LLMs) have shown clear advantages on solving complex problems such as mathematics, coding, and other domain-specific tasks~\cite{guo2025deepseek,cai2024internlm2,yang2025qwen3,wang2025internvl3,vteam2025glm45vglm41vthinkingversatilemultimodal,qin2026humansense}. 
DeepSeek-R1~\cite{guo2025deepseek} has pioneered a widely adopted reasoning-oriented post-training recipe, which first performs supervised fine-tuning (SFT) on cold-start Chain-of-Thought (CoT) data, followed by reinforcement learning (RL) with algorithms such as GRPO~\cite{shao2024deepseekmath}. However, in multimodal settings, it remains uncertain whether reasoning training can deliver consistent gains across diverse multimodal tasks. The trend of releasing parallel ``Instruct'' and ``Thinking'' models~\cite{bai2025qwen3vltechnicalreport,xu2025qwen3omnitechnicalreport} by leading teams is both resource-intensive and user-unfriendly. When evaluated across a broad suite of benchmarks, ``thinking'' models do not exhibit a universal advantage. 

The efficacy of reasoning-oriented post-training is influenced by multiple factors, including the characteristics of tasks, the capability of the base model, and the quality of CoT data~\cite{wang2025multimodal, li2025perception}. Prior studies find that for certain perception-oriented multimodal tasks, reasoning may introduce additional hallucinations~\cite{liu2025more, zheng2024thinking, wu2025combating, dong2025mirage} due to the increased amount of generated content, which in turn degrades model performance.  Consistent with these findings, we also observe similar limitations when enhancing the spatial understanding capabilities~\cite{zhu2025cvbench,yang2025thinking} of multimodal models through reasoning.

However, a principled criterion for determining when reasoning post-training is beneficial and which data should support it are still lacking. Concretely, given a target multimodal task and a candidate data source, we first need to assess the overall data efficacy, namely, whether direct fine-tuning on raw Visual Question-Answer data can yield measurable improvements over the base model. Building on this, we further need to evaluate whether the CoT data, constructed via expert annotation or model distillation, encodes useful thinking patterns that enable reasoning post-training to deliver superior gains. Moreover, since a multimodal task typically encompasses multiple subtypes with considerable variation across corresponding data samples, it is essential to identify and filter out data that are detrimental to model improvement from the full dataset. Addressing all these requirements demands prohibitively large experimental budgets.

In this paper, we introduce Dual Tuning, a data curation framework that quantifies training data efficacy and reasoning post-training suitability for target multimodal tasks and base models. The core idea is to sample task-specific training data to construct paired CoT and Direct-Answer (DA) examples, and jointly conduct SFT under their respective system prompts. By evaluating both the absolute gains of each training mode and the relative gains between them, Dual Tuning guides appropriate data composition and post-training strategy selection within a single experiment. By replacing unreliable human intuition with a principled comparative protocol, Dual Tuning accounts for variations in both data content and base models, and substantially reduces the experimental burden of preliminary ablation studies.

Our main contributions are threefold:

1) We propose Dual Tuning, a reasoning efficacy-driven data curation framework that jointly conducts reasoning and non-reasoning fine-tuning on paired CoT and Direct-Answer data, providing a systematic assessment of reasoning efficacy for Multimodal LLM training.

2) We apply Dual Tuning across diverse multimodal domains, including spatial reasoning, mathematics, and multi-disciplinary tasks, and further analyze how reinforcement training and thinking patterns affect reasoning efficacy, establishing comprehensive insights into the conditions under which reasoning post-training yields genuine benefits.

3) The proposed metrics guide data curation by identifying data that benefits from reasoning post-training, data better suited to direct-answer training, and data harmful to model improvement under either training mode, providing quantitative criteria for selecting appropriate training data and matching post-training strategies in Multimodal LLM training.

\section{Method}
Given a target multimodal task $\mathcal{T}$, a base model $\mathcal{M}$, and a candidate training dataset $\mathcal{D}$, Dual Tuning assesses reasoning efficacy by systematically comparing the gains of reasoning and non-reasoning training modes within a single fine-tuning experiment. Dual Tuning operates at subtask granularity. Each sample in $\mathcal{D}$ is annotated with a subtask label $c_k \in \mathcal{C}$, enabling fine-grained identification of which data groups are genuinely beneficial.


\subsection{Dual Tuning Algorithm}

We formalize Dual Tuning in Algorithm~\ref{alg:dual_tuning}. The paired datasets $\mathcal{D}_{\text{CoT}}$ and $\mathcal{D}_{\text{DA}}$, assigned respective system prompts, are jointly used to fine-tune $\mathcal{M}$ in a single training run, yielding the dual-tuned model $\mathcal{M}_{\text{DT}}$. Both $\mathcal{M}$ and $\mathcal{M}_{\text{DT}}$ are then evaluated under thinking and non-thinking prompts on the corresponding benchmark $\mathcal{B}$, producing scores $\{\mathbf{B_L, B_S}\}$ and $\{\mathbf{DT_L, DT_S}\}$ from which the efficacy metrics (Section~\ref{sec:metrics}) are computed to guide data curation decisions.

The Thinking mode uses following prompt: 
\begin{promptbox}
When the user asks a question, your response must 
include two parts: first, the reasoning process enclosed in <think>...</think> tags, then the final answer enclosed in <answer>...</answer> tags.
\end{promptbox}

While the Direct Answer mode uses:
\begin{promptbox}
Please directly answer the following question.
\end{promptbox}

\begin{algorithm}[t]
\caption{Dual Tuning Algorithm}
\label{alg:dual_tuning}
\begin{algorithmic}[1]
\Require Base model $\mathcal{M}$, Paired datasets $\mathcal{D}_{\text{CoT}}$, $\mathcal{D}_{\text{DA}}$, Subtasks $\mathcal{C} = \{c_1, \ldots, c_K\}$, Benchmark $\mathcal{B}$
\Ensure Efficacy metrics $\{ \mathbf{GAP_{B}}$, $\mathbf{GAP_{DT}}$, $\mathbf{Gain_{CoT}}$, $\mathbf{Gain_{DA}}$, $\mathbf{Gain_{token}} \}$ and data curation decisions
\Statex
\Statex \textit{// Stage 1: Data Preparation}
\State Construct paired data from $\mathcal{D}$, where $\mathcal{D}_{\text{CoT}}$ differs from $\mathcal{D}_{\text{DA}}$ only by the inclusion of CoT content, while the questions and answers remain the same.
\State Annotate each sample in $\mathcal{D}$ with subtask label $c_k \in \mathcal{C}$
\State Assign thinking/non-thinking system prompts to $\mathcal{D}_{\text{CoT}}$/$\mathcal{D}_{\text{DA}}$ respectively
\Statex
\Statex \textit{// Stage 2: Dual Fine-tuning (SFT)}
\State $\mathcal{M}_{\text{DT}} \leftarrow \text{SFT}(\mathcal{M}, \mathcal{D}_{\text{CoT}} \cup \mathcal{D}_{\text{DA}})$ \Comment{\textit{single joint training run}}
\Statex
\Statex \textit{// Stage 3: Efficacy Assessment}
\State $\mathbf{B_L}, \mathbf{B_S} \leftarrow \text{Eval}(\mathcal{M}, \mathcal{B}, \text{thinking/non-thinking prompt})$
\State $\mathbf{DT_L}, \mathbf{DT_S} \leftarrow$ $\text{Eval}(\mathcal{M}_{\text{DT}}, \mathcal{B}, \text{thinking/non-thinking prompt})$
\State Compute $\{ \mathbf{GAP_{B}}$, $\mathbf{GAP_{DT}}$, $\mathbf{Gain_{CoT}}$, $\mathbf{Gain_{DA}}$, $\mathbf{Gain_{token}} \}$ (Section~\ref{sec:metrics})
\State Derive data curation decisions (Section~\ref{sec:metrics})
\end{algorithmic}
\end{algorithm}

\subsection{Metrics}
\label{sec:metrics}
The tailored Dual Tuning metrics are as follows:
\begin{align}
  \mathbf{GAP_{B} = B_L - B_S}
\end{align}
represents the advantage of the base model in CoT evaluation ($\mathbf{B_L}$) relative to DA evaluation ($\mathbf{B_S}$).
\begin{align}
  \mathbf{GAP_{DT} = DT_L - DT_S}
\end{align}
represents the advantage of the Dual-Tuned model in CoT evaluation ($\mathbf{DT_L}$) relative to DA evaluation ($\mathbf{DT_S}$).
\begin{align}
  \mathbf{Gain_{CoT} = \frac{DT_L - \max(B_L, B_S)}{\max(B_L, B_S)}} \times 100
\end{align}
represents the gain achieved by CoT training relative to the base model's best performance.
\begin{align}
  \mathbf{Gain_{DA} = \frac{DT_S - \max(B_L, B_S)}{\max(B_L, B_S)}} \times 100
\end{align}
represents the real gain achieved by DA training relative to the base model's best performance.
\begin{align}
  \mathbf{Gain_{token}} = \frac{\mathbf{Gain_{CoT}}}{N_{token}} \times 100
\end{align}
represents the token-level gain for CoT training, where $N_{token}$ denotes the average number of output tokens in CoT evaluation.

Based on the above metrics and the subtask labels assigned to each data sample, the training data can be partitioned into three categories. A data subset is considered suitable for reasoning training only when both $\mathbf{Gain_{CoT}}$ and $\mathbf{GAP_{DT}}$ are positive. The data is better suited for DA training when $\mathbf{Gain_{DA}}$ is positive while $\mathbf{GAP_{DT}}$ is negative. When both $\mathbf{Gain_{CoT}}$ and $\mathbf{Gain_{DA}}$ are negative, the data should be filtered out.

\section{Experiments}

\subsection{Experimental Setup}
\textbf{Tasks and Benchmarks:}
We experiment on three well-studied multimodal reasoning tasks: spatial understanding~\cite{ouyang2025spacer, ma20253dsrbench, ai2025m2, jia2025omnispatial}, mathematics~\cite{qiao2025we, wang2024exploring}, and
multi-disciplinary question answering~\cite{su2025thinking}. For each task, we select representative benchmarks~\cite{yang2025thinking, zhu2025cvbench, lu2023mathvista, yue2024mmmu} for evaluation.

\textbf{Basemodel:}
We use Qwen2.5-VL-7B~\cite{bai2025qwen2} as the base model for its wide adoption in the community. To mitigate model and architecture-specific bias, we also experiment on Ming-lite-omni v1.5, a 20B Mixture-of-Experts (MoE) model with 3B active parameters~\cite{Mingomni2025}, with results shown in the Appendix.

\textbf{Datasets:}
We prepare datasets for both spatial and multi-disciplinary domains. Each data sample is generated in paired CoT and DA formats, ensuring that the question, visual input, and ground-truth answer remain identical across both versions.

\begin{itemize}[leftmargin=2.5em]
    \item Spatial Data: We construct spatial data from open-source 3D annotated datasets~\cite{roberts2021hypersim,dai2017scannet,yeshwanth2023scannet++,baruch2021arkitscenes,ray2024sat}. Building on the original Visual Question-Answer(VQA) data, we use Gemini 2.5 Pro~\cite{comanici2025gemini} to construct additional chain-of-thought (CoT) content. The final dataset is balanced across eight categories of VSI-Bench~\cite{su2025thinking} and four categories of CV-Bench~\cite{zhu2025cvbench}. For SFT, the dataset comprises 30k image-text VQA and 8k video-text VQA pairs. Additionally, a separate set of 10k image-text and 10k video-text VQA samples is reserved for RL stages.

    \item Multi-disciplinary Data: we test the open-source OneThinker~\cite{feng2025onethinker} dataset in math and multi-disciplinary experiments. To ensure domain relevance, we exclude data categories weakly correlated with MMMU and MathVista~\cite{lu2023mathvista,yue2024mmmu}, such as tracking and segmentation. We remove samples in which "wait" appears more than twice in the CoT content to prevent the model from falling into unproductive reflection. We sample 20k image-text VQA pairs for SFT and 10k pairs for RL.
    
\end{itemize}

\textbf{Implementation Details:}
All SFT and RL experiments are conducted on a cluster of 32 NVIDIA H800 GPUs for 4 epochs using the AdamW optimizer. For SFT, the learning rate is set to $1 \times 10^{-5}$, with a per-GPU batch size of 4 for images and 1 for videos. The maximum video frame count is set to 256. For RL, we employ Group Relative Policy Optimization (GRPO) with a group size of 8 and a KL coefficient of 0.01. The RL stage use a learning rate of $1 \times 10^{-6}$, a per-GPU batch size of 1, and a reduced maximum video frame count of 64 for computational efficiency.

\begin{table}
  \caption{Results of preliminary experiments on spatial tasks. \textbf{Baseline}: Qwen2.5-VL-7B. \textbf{I}: Image Spatial Data. \textbf{V}: Video Spatial Data. \textbf{S}: Direct-Answer. \textbf{L}: CoT.}
  \label{tab:t1}
  \centering
  \setlength{\tabcolsep}{10 pt}
  \small
  \begin{tabular}{l|cc|cc}
  \toprule
  \multirow{2}{*}{Model} & \multicolumn{2}{c|}{VSI (Vid)} & \multicolumn{2}{c}{CV (Img)} \\
          & Eval-S & Eval-L & Eval-S & Eval-L \\
  \midrule
  Baseline & 39.40 & 26.78 & 75.17 & 73.81 \\
  \midrule
  E1: I-S & \textcolor{gray}{37.94} & \textcolor{gray}{37.34} &	\textcolor{black}{84.15} &	\textcolor{gray}{84.27} \\
  E2: I-L &	\textcolor{gray}{12.96}	& \textcolor{gray}{37.38} &	\textcolor{gray}{75.51} &	\textcolor{black}{80.55} \\
  E3: V-S &	\textcolor{black}{50.98}	& \textcolor{gray}{50.50} &	\textcolor{gray}{78.20} &	\textcolor{gray}{77.82} \\
  E4: V-L & \textcolor{gray}{0.00} & \textcolor{black}{44.14} &	\textcolor{gray}{37.72} &	\textcolor{gray}{73.09} \\
  \midrule
  E5: IV-SL & 51.35	& 43.26 &	84.00 &	80.25 \\
 
  \bottomrule
\end{tabular}
\end{table}

\begin{table}
  \caption{Results of preliminary experiments on disciplinary tasks. \textbf{Baseline}: Qwen2.5-VL-7B. \textbf{O}: Onethinker Image Data. \textbf{S}: Direct-Answer. \textbf{L}: CoT.}
  \label{tab:t2}
  \centering
  \setlength{\tabcolsep}{10 pt}
  \small
  \begin{tabular}{l|cc|cc}
  \toprule
  \multirow{2}{*}{Model} & \multicolumn{2}{c|}{MathVista} & \multicolumn{2}{c}{MMMU} \\
          & Eval-S & Eval-L & Eval-S & Eval-L \\
  \midrule
  Baseline & 69.80 & 69.83 & 53.33 & 51.56 \\
  \midrule
  E6: O-S & \textcolor{black}{67.43} & \textcolor{gray}{67.43} &	\textcolor{black}{52.00} &	\textcolor{gray}{51.00} \\
  E7: O-L &	\textcolor{gray}{59.13}	& \textcolor{black}{73.40} &	\textcolor{gray}{25.89} &	\textcolor{black}{53.33} \\
  \midrule
  E8: O-SL & 67.37	& 72.87 &	51.00 &	54.11 \\
 
  \bottomrule
\end{tabular}
\end{table}

\begin{table*}[t]
  \caption{Dual Tuning results on spatial tasks. Values in red and green denote negative and positive results, respectively. A task is identified as suitable for reasoning-oriented training only when both $\mathbf{Gain_{CoT}}$ and $\mathbf{GAP_{DT}}$ exhibit concurrent positive values (highlighted in \textbf{\textcolor{myTextGreenColor}{green}})}
  \label{tab:spatial}
  \setlength{\tabcolsep}{4.8 mm}
  \centering
  \small
  \begin{tabular*}{0.98\linewidth}{l|ccc|ccccc}
  \toprule
  Task & $\mathbf{B_S}$ & $\mathbf{B_L}$ & $\mathbf{GAP_{B}}$ & $\mathbf{DT_S}$ & $\mathbf{DT_L}$ & $\mathbf{Gain_{DA}}$ & \textcolor{myTextGreenColor}{$\mathbf{Gain_{CoT}}$} & \textcolor{myTextGreenColor}{$\mathbf{GAP_{DT}}$} \\
  \midrule
  \textcolor{myTextBenchColor}{VSIBench} & 37.57 & 24.91 & \ratioCell{-12.66} & 51.35 & 43.26 & \ratioCell{36.68} & \ratioCell{15.15} & \ratioCell{-8.09} \\
  \midrule
  \cellcolor{orange!10} Obj. Count & 40.41 & 12.19 & \ratioCell{-28.22} & 49.65 & 43.47 & \ratioCell{22.87} & \ratioCell{7.57} & \ratioCell{-6.18} \\
  \cellcolor{orange!10} Abs. Dist. & 22.88 & 1.80 & \ratioCell{-21.08} & 40.17 & 29.68 & \ratioCell{75.57} & \ratioCell{29.72} & \ratioCell{-10.49} \\
  \cellcolor{orange!10} Obj. Size & 50.18 & 12.34 & \ratioCell{-37.84} & 64.30 & 50.97 & \ratioCell{28.14} & \ratioCell{1.57} & \ratioCell{-13.33} \\
  \cellcolor{orange!10} Room Size & 38.96 & 22.71 & \ratioCell{-16.25} & 56.98 & 38.72 & \ratioCell{46.25} & \ratioCell{-0.62} & \ratioCell{-18.26} \\
  \cellcolor{orange!10} Rel. Dist. & 39.44 & 40.14 & \ratioCell{0.70} & 58.31 & 48.03 & \ratioCell{45.27} & \ratioCell{19.66} & \ratioCell{-10.28} \\
  \cellcolor{orange!10} Rel. Dir. & 39.66 & 39.57 & \ratioCell{-0.09} & 44.44 & 43.82 & \ratioCell{12.05} & \ratioCell{10.49} & \ratioCell{-0.62} \\
  \cellcolor{orange!10} Route Plan & 32.47 & 32.47 & \ratioCell{0.00} & 34.02 & 32.47 & \ratioCell{4.77} & \ratioCell{0.00} & \ratioCell{-1.55} \\
  \cellcolor{orange!10} Appr. Order & 36.57 & 38.03 & \ratioCell{1.46} & 62.94 & 58.90 & \ratioCell{65.50} & \ratioCell{54.88} & \ratioCell{-4.04} \\
  \midrule
  \textcolor{myTextBenchColor}{CVBench} & 75.17 & 73.81 & \ratioCell{-1.36} & 84.00 & 80.25 & \ratioCell{11.75} & \ratioCell{6.76} & \ratioCell{-3.75} \\
  \midrule
  \cellcolor{cyan!10} Count & 64.72 & 62.94 & \ratioCell{-1.78} & 70.94 & 61.80 & \ratioCell{9.61} & \ratioCell{-4.51} & \ratioCell{-9.14} \\
  \cellcolor{cyan!10} Relation & 87.69 & 86.46 & \ratioCell{-1.23} & 95.69 & 92.92 & \ratioCell{9.12} & \ratioCell{5.96} & \ratioCell{-2.77} \\
  \cellcolor{cyan!10} Depth & 70.17 & 72.00 & \ratioCell{1.83} & 87.00 & 87.00 & \ratioCell{20.83} & \ratioCell{20.83} & \ratioCell{0.00} \\
  \cellcolor{cyan!10} Distance & 80.33 & 76.17 & \ratioCell{-4.16} & 85.50 & 84.00 & \ratioCell{6.44} & \ratioCell{4.57} & \ratioCell{-1.50} \\
  \bottomrule
\end{tabular*}
\end{table*}

\begin{table*}[t]
  \caption{Dual Tuning results on MathVista tasks. Values in red and green denote negative and positive results. A task is identified as suitable for reasoning-oriented training only when both $\mathbf{Gain_{CoT}}$ and $\mathbf{GAP_{DT}}$ exhibit concurrent positive values (highlighted in \textbf{\textcolor{myTextGreenColor}{green}}).}
  \label{tab:mathvista}
  \small
  \setlength{\tabcolsep}{5.0 mm}
  \centering
  \begin{tabular*}{0.98\linewidth}{l|ccc|ccccc}
  \toprule
  Task & $\mathbf{B_S}$ & $\mathbf{B_L}$ & $\mathbf{GAP_{B}}$ & $\mathbf{DT_S}$ & $\mathbf{DT_L}$ & $\mathbf{Gain_{DA}}$ & \textcolor{myTextGreenColor}{$\mathbf{Gain_{CoT}}$} & \textcolor{myTextGreenColor}{$\mathbf{GAP_{DT}}$} \\
  \midrule
  \textcolor{myTextBenchColor}{MathVista} & 69.80 & 69.83 & \ratioCell{0.03} & 67.37 & 72.87 & \ratioCell{-3.52} & \ratioCell{4.35} & \ratioCell{5.50} \\
  \midrule
  \cellcolor{yellow!10} Comm. & 44.44 & 40.51 & \ratioCell{-3.94} & 49.07 & 46.06 & \ratioCell{10.42} & \ratioCell{3.65} & \ratioCell{-3.01} \\
  \cellcolor{yellow!10} Scientific & 74.70 & 74.40 & \ratioCell{-0.30} & 74.70 & 74.70 & \ratioCell{0.00} & \ratioCell{0.00} & \ratioCell{0.00} \\
  \cellcolor{yellow!10} Geometry & 70.55 & 70.26 & \ratioCell{-0.29} & 64.08 & 74.57 & \ratioCell{-9.17} & \ratioCell{5.70} & \ratioCell{10.49} \\
  \cellcolor{yellow!10} Arithmetic & 78.33 & 80.45 & \ratioCell{2.12} & 71.52 & 82.12 & \ratioCell{-11.10} & \ratioCell{2.08} & \ratioCell{10.60} \\
  \cellcolor{yellow!10} Logical & 23.42 & 28.83 & \ratioCell{5.41} & 18.92 & 23.42 & \ratioCell{-34.37} & \ratioCell{-18.77} & \ratioCell{4.50} \\
  \cellcolor{yellow!10} Statistical & 84.64 & 84.12 & \ratioCell{-0.52} & 85.86 & 87.43 & \ratioCell{1.44} & \ratioCell{3.30} & \ratioCell{1.57} \\
  \cellcolor{yellow!10} Algebraic & 68.75 & 70.83 & \ratioCell{2.08} & 66.15 & 77.08 & \ratioCell{-6.61} & \ratioCell{8.82} & \ratioCell{10.93} \\

  \bottomrule
\end{tabular*}
\end{table*}

\begin{table*}[t]
  \caption{Experimental results on MMMU tasks. Values in red and green denote negative and positive results, respectively. A task is identified as suitable for reasoning-oriented training only when both $\mathbf{Gain_{CoT}}$ and $\mathbf{GAP_{DT}}$ exhibit concurrent positive values (highlighted in \textbf{\textcolor{myTextGreenColor}{green}})}
  \label{tab:mmmu}
  \setlength{\tabcolsep}{4.5 mm}
  \centering
  \small
  \begin{tabular*}{0.98\linewidth}{l|ccc|ccccc}
  \toprule
  Task & $\mathbf{B_S}$ & $\mathbf{B_L}$ & $\mathbf{GAP_{B}}$ & $\mathbf{DT_S}$ & $\mathbf{DT_L}$ & $\mathbf{Gain_{DA}}$ & \textcolor{myTextGreenColor}{$\mathbf{Gain_{CoT}}$} & \textcolor{myTextGreenColor}{$\mathbf{GAP_{DT}}$} \\
  \midrule
  \textcolor{myTextBenchColor}{MMMU} & 53.33 & 51.56 & \ratioCell{-1.77} & 51 & 54.11 & \ratioCell{-4.37} & \ratioCell{1.46} & \ratioCell{3.11} \\
  \midrule
  \cellcolor{blue!10} Art & 70.00 & 70.00 & \ratioCell{0.00} & 73.33 & 73.33 & \ratioCell{4.76} & \ratioCell{4.76} & \ratioCell{0.00} \\
  \cellcolor{blue!10} Design & 76.67 & 76.67 & \ratioCell{0.00} & 63.33 & 73.33 & \ratioCell{-17.40} & \ratioCell{-4.36} & \ratioCell{10.00} \\
  \cellcolor{blue!10} Music & 30.00 & 26.67 & \ratioCell{-3.33} & 43.33 & 30.00 & \ratioCell{44.43} & \ratioCell{0.00} & \ratioCell{-13.33} \\
  \cellcolor{blue!10} Art Theory & 83.33 & 76.67 & \ratioCell{-6.66} & 80.00 & 86.67 & \ratioCell{-4.00} & \ratioCell{4.01} & \ratioCell{6.67} \\
  \cellcolor{red!10} Accounting & 56.67 & 66.67 & \ratioCell{10.00} & 36.67 & 53.33 & \ratioCell{-45.00} & \ratioCell{-20.01} & \ratioCell{16.66} \\
  \cellcolor{red!10} Finance & 40.00 & 43.33 & \ratioCell{3.33} & 36.67 & 43.33 & \ratioCell{-15.37} & \ratioCell{0.00} & \ratioCell{6.66} \\
  \cellcolor{red!10} Economics & 56.67 & 53.33 & \ratioCell{-3.34} & 43.33 & 60.00 & \ratioCell{-23.54} & \ratioCell{5.88} & \ratioCell{16.67} \\
  \cellcolor{red!10} Manage & 43.33 & 36.67 & \ratioCell{-6.66} & 46.67 & 46.67 & \ratioCell{7.71} & \ratioCell{7.71} & \ratioCell{0.00} \\
  \cellcolor{red!10} Marketing & 63.33 & 60.00 & \ratioCell{-3.33} & 56.67 & 63.33 & \ratioCell{-10.52} & \ratioCell{0.00} & \ratioCell{6.66} \\
  \cellcolor{yellow!10} Math & 46.67 & 46.67 & \ratioCell{0.00} & 46.67 & 56.67 & \ratioCell{0.00} & \ratioCell{21.43} & \ratioCell{10.00} \\
  \cellcolor{yellow!10} Biology & 36.67 & 40.00 & \ratioCell{3.33} & 36.67 & 43.33 & \ratioCell{-8.33} & \ratioCell{8.33} & \ratioCell{6.66} \\
  \cellcolor{yellow!10} Chemistry & 33.33 & 43.33 & \ratioCell{10.00} & 36.67 & 40.00 & \ratioCell{-15.37} & \ratioCell{-7.69} & \ratioCell{3.33} \\
  \cellcolor{yellow!10} Geography & 43.33 & 40.00 & \ratioCell{-3.33} & 50.00 & 30.00 & \ratioCell{15.39} & \ratioCell{-30.76} & \ratioCell{-20.00} \\
  \cellcolor{yellow!10} Physics & 33.33 & 40.00 & \ratioCell{6.67} & 36.67 & 53.33 & \ratioCell{-8.33} & \ratioCell{33.33} & \ratioCell{16.66} \\
  \cellcolor{cyan!10} Basic Med. Sci. & 60.00 & 56.67 & \ratioCell{-3.33} & 63.33 & 66.67 & \ratioCell{5.55} & \ratioCell{11.12} & \ratioCell{3.34} \\
  \cellcolor{cyan!10} Clinical Med. & 60.00 & 70.00 & \ratioCell{10.00} & 53.33 & 66.67 & \ratioCell{-23.81} & \ratioCell{-4.76} & \ratioCell{13.34} \\
  \cellcolor{cyan!10} Diag. \& Lab. Med. & 46.67 & 43.33 & \ratioCell{-3.34} & 33.33 & 33.33 & \ratioCell{-28.58} & \ratioCell{-28.58} & \ratioCell{0.00} \\
  \cellcolor{cyan!10} Public Health & 70.00 & 66.67 & \ratioCell{-3.33} & 60.00 & 66.67 & \ratioCell{-14.29} & \ratioCell{-4.76} & \ratioCell{6.67} \\
  \cellcolor{cyan!10} Pharmacy & 73.33 & 56.67 & \ratioCell{-16.66} & 56.67 & 66.67 & \ratioCell{-22.72} & \ratioCell{-9.08} & \ratioCell{10.00} \\
  \cellcolor{green!10} History & 73.33 & 60.00 & \ratioCell{-13.33} & 70.00 & 73.33 & \ratioCell{-4.54} & \ratioCell{0.00} & \ratioCell{3.33} \\
  \cellcolor{green!10} Literature & 86.67 & 80.00 & \ratioCell{-6.67} & 86.67 & 80.00 & \ratioCell{0.00} & \ratioCell{-7.70} & \ratioCell{-6.67} \\
  \cellcolor{green!10} Psychology & 66.67 & 60.00 & \ratioCell{-6.67} & 60.00 & 76.67 & \ratioCell{-10.00} & \ratioCell{15.00} & \ratioCell{16.67} \\
  \cellcolor{green!10} Sociology & 53.33 & 50.00 & \ratioCell{-3.33} & 63.33 & 66.67 & \ratioCell{18.75} & \ratioCell{25.01} & \ratioCell{3.34} \\
  \cellcolor{orange!10} Agriculture & 50.00 & 50.00 & \ratioCell{0.00} & 63.33 & 50.00 & \ratioCell{26.66} & \ratioCell{0.00} & \ratioCell{-13.33} \\
  \cellcolor{orange!10} Arch. \& Eng. & 36.67 & 33.33 & \ratioCell{-3.34} & 36.67 & 33.33 & \ratioCell{0.00} & \ratioCell{-9.11} & \ratioCell{-3.34} \\
  \cellcolor{orange!10} Comp. Sci. & 60.00 & 56.67 & \ratioCell{-3.33} & 50.00 & 50.00 & \ratioCell{-16.67} & \ratioCell{-16.67} & \ratioCell{0.00} \\
  \cellcolor{orange!10} Electronics & 33.33 & 43.33 & \ratioCell{10.00} & 33.33 & 36.67 & \ratioCell{-23.08} & \ratioCell{-15.37} & \ratioCell{3.34} \\
  \cellcolor{orange!10} Energy \& Power & 43.33 & 36.67 & \ratioCell{-6.66} & 53.33 & 43.33 & \ratioCell{23.08} & \ratioCell{0.00} & \ratioCell{-10.00} \\
  \cellcolor{orange!10} Materials & 23.33 & 30.00 & \ratioCell{6.67} & 30.00 & 36.67 & \ratioCell{0.00} & \ratioCell{22.23} & \ratioCell{6.67} \\
  \cellcolor{orange!10} Mech. Eng. & 50.00 & 33.33 & \ratioCell{-16.67} & 30.00 & 23.33 & \ratioCell{-40.00} & \ratioCell{-53.34} & \ratioCell{-6.67} \\

  \bottomrule
\end{tabular*}
\end{table*}

\subsection{Preliminary Validation of the Dual Tuning }

We conduct preliminary experiments by comparing Dual Tuning with single-mode fine-tuning baselines to validate that its joint training design can recover the conclusions of separate ablation studies.

In spatial tasks (Table~\ref{tab:t1}), joint training (E5) on both data types and modalities largely preserves the peak performance achieved by the corresponding single-mode baselines, with deviations of less than 1 percentage point. These results support the validity of Dual Tuning as a means of capturing the gains of both CoT and DA training. Moreover, training exclusively on CoT data causes outputs in DA evaluation to become verbose, disrupting the evaluation on VSI-Bench (E2/E4). Conversely, DA training causes the CoT inference to degenerate into direct answers, where the score actually reflects DA capability (E1/E3). Comparing independent video and image training, we find that image data exerts a more significant influence on video evaluations than vice versa. On the overall metrics, both CoT and DA training outperform the base model in the spatial domain. However, the gains from DA training are substantially larger than those from CoT training.

For multi-disciplinary tasks, we observe that Dual Tuning consistently matches the peak performance achieved by single-mode fine-tuning. As shown in Table~\ref{tab:t2}, the cross-mode effects in single-mode training (E6/E7) are mitigated because the GPT-assisted evaluation used in benchmarks is less sensitive to format. Notably, CoT training outperforms DA training in both mathematical and multi-disciplinary tasks, leading to substantial positive gains. This pattern contrasts sharply with that observed in spatial tasks.

\subsection{Dual Tuning on Spatial Tasks}

We evaluate the Dual Tuning results on spatial tasks at a fine-grained subtask level (Table~\ref{tab:spatial}). According the initial performance of the Qwen2.5-VL-7B in subtasks like \textit{Object Count}, \textit{Absolute Distance}, \textit{Absolute Size}, and \textit{Room Size}, negative $\mathbf{GAP_{B}}$ means it performs better in the Direct-Answer (DA) mode than Chain-of-Though (CoT). For \textit{Relative Distance}, \textit{Relative Direction}, and \textit{Route Plan} subtasks, no significant difference between CoT and DA inference, indicating that the initial capabilities are comparable. \textit{Appearance Order} is the only subtask where the base model shows a marginal positive gain under CoT inference. The CV-Bench results point to a consistent conclusion that the current base model is better suited to DA-mode inference on spatial tasks.

After Dual Tuning, both DA and CoT evaluation achieve positive gains over the base model, evidenced by positive values for both $\mathbf{Gain_{CoT}}$ and $\mathbf{Gain_{DA}}$. However, no subtask satisfies the criterion for reasoning superiority, where both $\mathbf{GAP_{DT}}$ and $\mathbf{Gain_{CoT}}$ being positive. Red $\mathbf{GAP_{DT}}$ reveals that the improvement driven by DA is significantly superior to that of CoT, approaching parity only in tasks such as \textit{Relative Direction} and \textit{Route Plan}. 

Results with Ming-lite-omni are presented in Table~\ref{tab:ming-sp} (Appendix). The $\mathbf{GAP_{B}}$ varies from Qwen2.5-VL and shows less contrast between inference modes. The Dual Tuning results  ($\mathbf{GAP_{DT}}$) are highly consistent with Qwen2.5-VL,  the negative value indicates that DA training yields more improvements on these spatial tasks. The above results suggest that under the current base model and data conditions, DA training is the better choice than reasoning-oriented supervision for improving spatial capabilities related to VSI-Bench and CV-Bench.

\begin{figure*}[!ht]
  \vspace{0.5cm}
  \centering
  \includegraphics[width=0.99\linewidth]{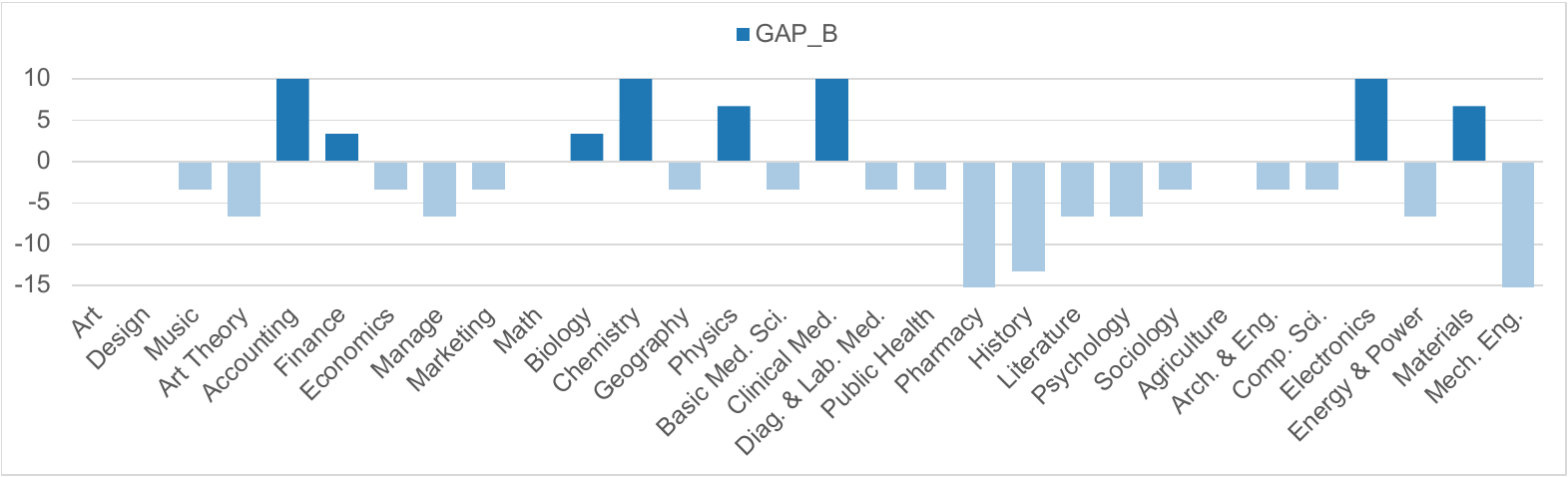}
  \caption{On multi-disciplinary tasks, the base model exhibits substantial variation in its initial performance between CoT and DA inference across different subtasks. Positive values indicate an advantage of CoT inference.}
  \label{fig:mmmu_gap_b}
\end{figure*}

\begin{figure*}[!ht]
  \centering
  \includegraphics[width=0.99\linewidth]{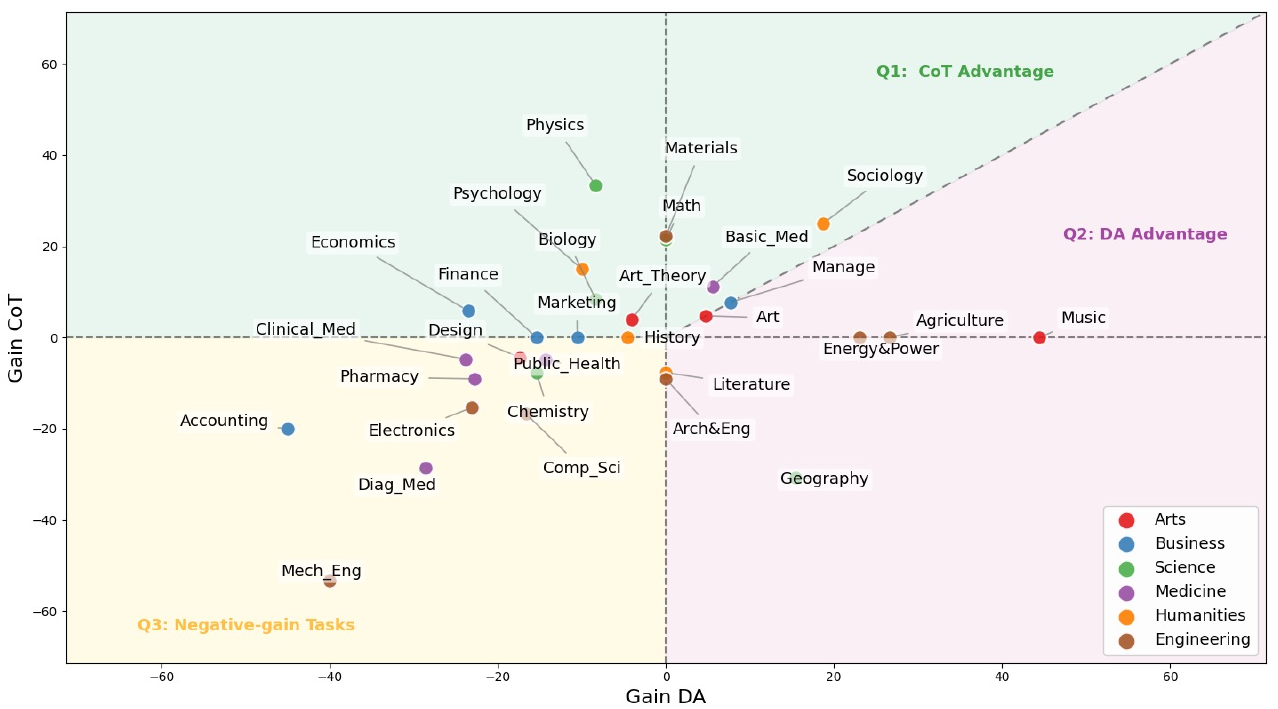}
  \caption{We plot each task's $\mathbf{Gain_{CoT}}$ and $\mathbf{Gain_{DA}}$ in a two-dimensional coordinate map. Through three distinct regions, we categorize the suitability of different tasks for the two training modes.}
  \label{fig:mmmu_sca}
  \vspace{0.5cm}
\end{figure*}

\begin{table*}[!ht]
  \caption{Adding RL Training on Dual-Tuned Models for Spatial Tasks.  A task is identified as suitable for reasoning-oriented training only when both $\mathbf{Gain_{CoT}}$ and $\mathbf{GAP_{DT}}$ exhibit concurrent positive values.}
  \label{tab:sp-rl}
  \setlength{\tabcolsep}{7 mm}
  \centering
  \small
  \begin{tabular*}{0.98\linewidth}{l|ccc|ccccc}
  \toprule
  Task & $\mathbf{Gain_{DA}}$ & \textcolor{myTextGreenColor}{$\mathbf{Gain_{CoT}}$} & \textcolor{myTextGreenColor}{$\mathbf{GAP_{DT}}$} & $\mathbf{Gain_{DA}^{RL}}$ & \textcolor{myTextGreenColor}{$\mathbf{Gain_{CoT}^{RL}}$} & \textcolor{myTextGreenColor}{$\mathbf{GAP_{DT}^{RL}}$} \\
  \midrule
  \textcolor{myTextBenchColor}{VSIBench} & \ratioCell{36.68} & \ratioCell{15.15} & \ratioCell{-8.09} & \ratioCell{36.81} & \ratioCell{32.50} & \ratioCell{-1.62} \\
  \midrule
  \cellcolor{orange!10} Obj. Count & \ratioCell{22.87} & \ratioCell{7.57} & \ratioCell{-6.18} & \ratioCell{23.29} & \ratioCell{29.82} & \ratioCell{2.64} \\
  \cellcolor{orange!10} Abs. Dist. & \ratioCell{75.57} & \ratioCell{29.72} & \ratioCell{-10.49} & \ratioCell{76.75} & \ratioCell{75.31} & \ratioCell{-0.33} \\
  \cellcolor{orange!10} Obj. Size & \ratioCell{28.14} & \ratioCell{1.57} & \ratioCell{-13.33} & \ratioCell{28.36} & \ratioCell{28.50} & \ratioCell{0.07} \\
  \cellcolor{orange!10} Room Size & \ratioCell{46.25} & \ratioCell{-0.62} & \ratioCell{-18.26} & \ratioCell{40.81} & \ratioCell{13.91} & \ratioCell{-10.48} \\
  \cellcolor{orange!10} Rel. Dist. & \ratioCell{45.27} & \ratioCell{19.66} & \ratioCell{-10.28} & \ratioCell{47.01} & \ratioCell{33.68} & \ratioCell{-5.35} \\
  \cellcolor{orange!10} Rel. Dir. & \ratioCell{12.05} & \ratioCell{10.49} & \ratioCell{-0.62} & \ratioCell{12.48} & \ratioCell{10.77} & \ratioCell{-0.68} \\
  \cellcolor{orange!10} Route Plan & \ratioCell{4.77} & \ratioCell{0.00} & \ratioCell{-1.55} & \ratioCell{9.55} & \ratioCell{22.24} & \ratioCell{4.12} \\
  \cellcolor{orange!10} Appr. Order & \ratioCell{65.50} & \ratioCell{54.88} & \ratioCell{-4.04} & \ratioCell{64.24} & \ratioCell{56.59} & \ratioCell{-2.91} \\
  \midrule
  \textcolor{myTextBenchColor}{CVBench} & \ratioCell{11.75} & \ratioCell{6.76} & \ratioCell{-3.75} & \ratioCell{11.85} & \ratioCell{11.75} & \ratioCell{-0.08} \\
  \midrule
  \cellcolor{cyan!10} Count & \ratioCell{9.61} & \ratioCell{-4.51} & \ratioCell{-9.14} & \ratioCell{8.24} & \ratioCell{7.26} & \ratioCell{-0.63} \\
  \cellcolor{cyan!10} Relation & \ratioCell{9.12} & \ratioCell{5.96} & \ratioCell{-2.77} & \ratioCell{9.31} & \ratioCell{8.60} & \ratioCell{-0.62} \\
  \cellcolor{cyan!10} Depth & \ratioCell{20.83} & \ratioCell{20.83} & \ratioCell{0.00} & \ratioCell{21.99} & \ratioCell{23.61} & \ratioCell{1.17} \\
  \cellcolor{cyan!10} Distance & \ratioCell{6.44} & \ratioCell{4.57} & \ratioCell{-1.50} & \ratioCell{7.06} & \ratioCell{7.06} & \ratioCell{0.00} \\
  \bottomrule
\end{tabular*}
\end{table*}

\begin{figure*}[!ht]
  \centering
  \begin{tabular}{cc}
    \includegraphics[width=0.48\textwidth, valign=c]{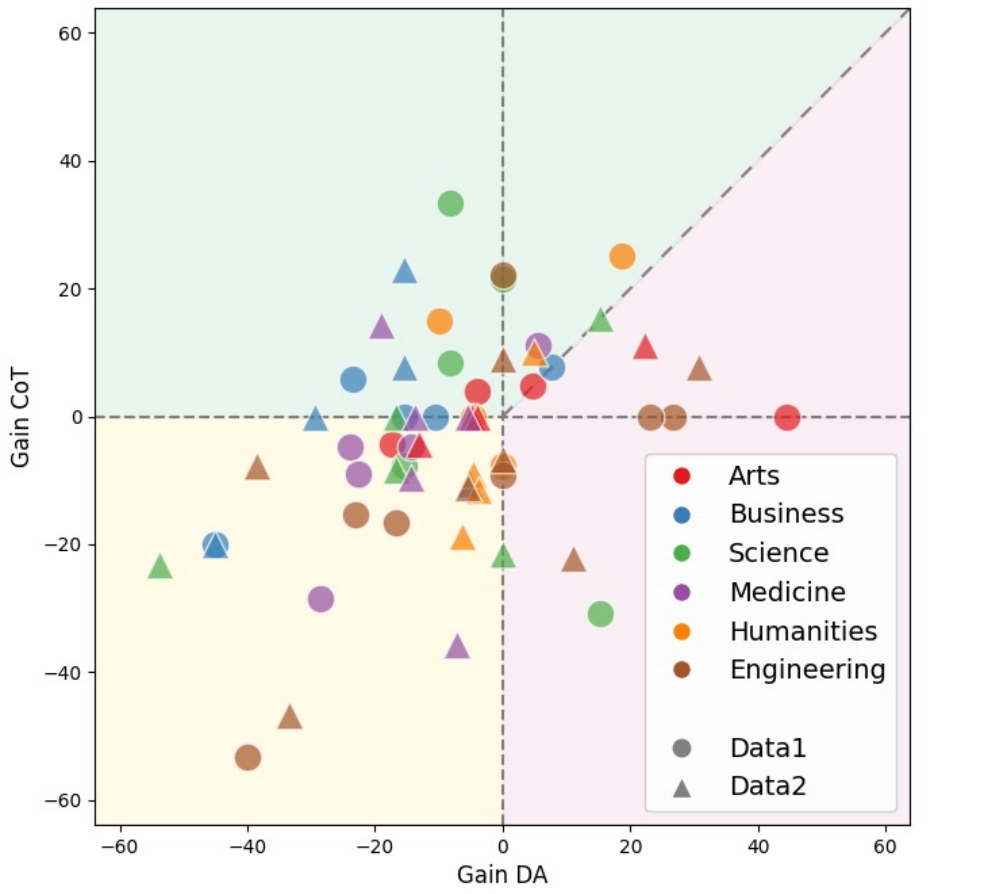} & 
    \includegraphics[width=0.48\textwidth, valign=c]{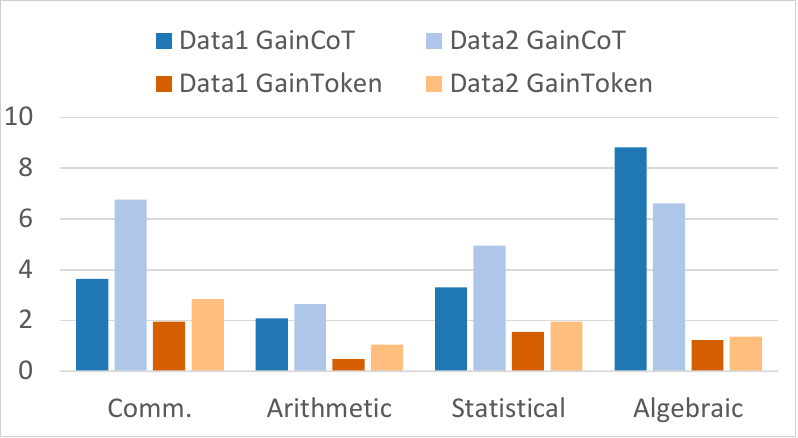} \\
    
    \begin{subfigure}[t]{0.48\textwidth}
      \caption{The original and new datasets are marked by circles and triangles. Different CoT content leads to markedly different distributions across subtasks.}
      \label{fig:think_pattern}
    \end{subfigure} & 
    \begin{subfigure}[t]{0.48\textwidth}
      \caption{The comparison of $\mathbf{Gain_{CoT}}$ and $\mathbf{Gain_{token}}$ between the two datasets is not entirely consistent, with opposite trends observed on algebraic subtasks.}
      \label{fig:token}
    \end{subfigure}
  \end{tabular}
  \caption{Impact of CoT Thinking Patterns on Data Efficacy (left) and Token-Level Gains (right)}
\end{figure*}

\begin{figure*}[!ht]
  \centering
  \includegraphics[width=0.99\linewidth]{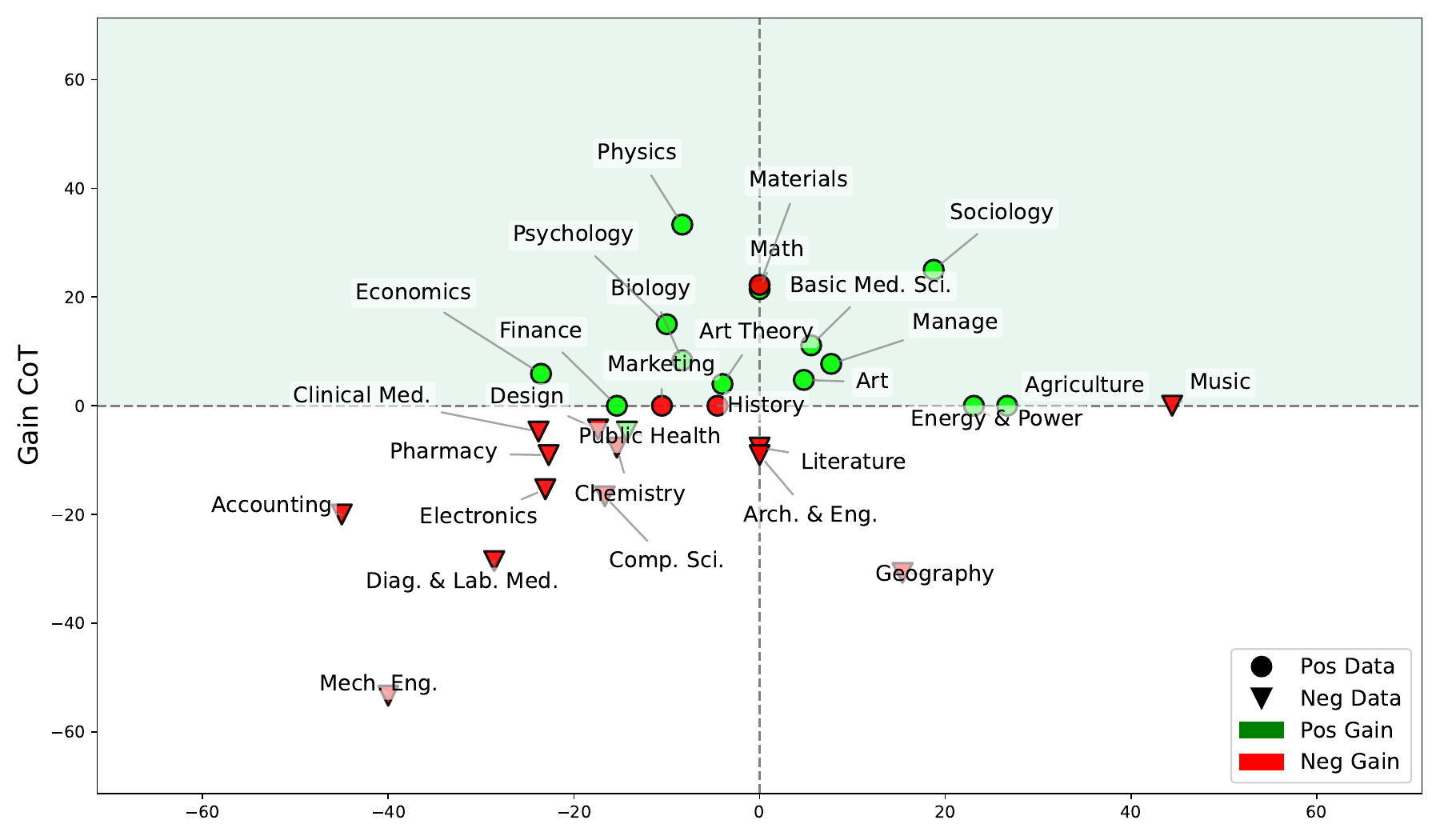}
  \caption{We partition tasks into two halves using $\mathbf{Gain_{CoT}}$ and conduct two separate CoT training on the data belonging to each half. The results show that the data corresponding to negative subtasks yield negative gains during standalone training, and vice versa.}
  \label{fig:sca_L}
\end{figure*}

\subsection{Dual Tuning on Mathematical Tasks}

The mathematical results exhibit the opposite pattern to those on spatial tasks (Table~\ref{tab:mathvista}). The Qwen2.5-VL shows similar base performance under CoT and DA prompting. Only \textit{Logical Reasoning} shows an advantage for CoT, while \textit{Numeric Commonsense} shows a preference for DA, consistent with intuition. After Dual Tuning,  green $\mathbf{Gain_{CoT}}$ indicates that most tasks achieve positive gains through CoT training, while DA training yields negative gains. The positive $\mathbf{GAP_{DT}}$ indicate that reasoning training with the current CoT data is more effective than DA training across categories except \textit{Numeric Commonsense}.

The results of Ming-lite-omni are presented in Table~\ref{tab:ming-mathvista} (Appendix). The $\mathbf{GAP_{B}}$ shows that Ming-lite-omni differs from Qwen2.5-VL, starting with a significant advantage in CoT inference. This high baseline causes $\mathbf{Gain_{CoT}}$ to turn negative for some subtasks. Combined with $\mathbf{GAP_{DT}}$, reasoning gains only appear in \textit{Geometric Reasoning}, \textit{Logical Reasoning}, and \textit{Statistical Reasoning}. These results indicate that reasoning efficacy on a given task is jointly influenced by both base model capabilities and data quality.

\subsection{Dual Tuning on Multi-disciplinary Tasks}

 Recent technical reports~\cite{vteam2025glm45vglm41vthinkingversatilemultimodal,bai2025qwen3vltechnicalreport} often categorize MathVista and MMMU within the same STEM~(Science, Technology, Engineering, and Mathematics) domain. But we observe more diverse patterns when analyzing the Dual Tuning results on MMMU. Detailed experimental results are provided in Table~\ref{tab:mmmu}.

Figure~\ref{fig:mmmu_gap_b} shows large variations in the base model's initial capability across different tasks. In tasks such as \textit{Accounting}, \textit{Chemistry}, and \textit{Physics}, the base model shows advantages in CoT, while in \textit{History} and \textit{Pharmacy} the DA mode perform better. Results with Ming-lite-omni as the base model are presented in Table~\ref{tab:ming-mmmu} (Appendix). $\mathbf{GAP_{B}}$ reveals a larger gap in initial capabilities across subtasks compared to Qwen2.5-VL. Notably, comparing $\mathbf{GAP_{DT}}$ and $\mathbf{Gain_{CoT}}$, the subtasks that exhibit reasoning gains reveal partially overlap with those in Qwen2.5-VL, suggesting that the data has a strong influence on reasoning efficacy.

To provide a more intuitive comparison across different subtasks, we plot a 2D map to visualize the $\mathbf{Gain_{CoT}}$ and $\mathbf{Gain_{DA}}$. Figure~\ref{fig:mmmu_sca} allows us to divide tasks into three regions according to their efficacy for two training modes. On tasks such as \textit{Math}, \textit{Physics}, \textit{Psychology}, \textit{Sociology}, and \textit{Basic Medical Science}, both $\mathbf{GAP_{DT}}$ and $\mathbf{Gain_{CoT}}$ are positive, suggesting that current data are better suited for CoT training. Conversely, for \textit{Music}, \textit{Geography}, and \textit{Agriculture}, $\mathbf{GAP_{DT}}$ is negative while $\mathbf{Gain_{DA}}$ is positive, indicating that DA training is more appropriate for such tasks. For \textit{Accounting}, \textit{Diagnostics and Laboratory Medicine}, \textit{Mechanical Engineering}, and \textit{Public Health}, both $\mathbf{Gain_{CoT}}$ and $\mathbf{Gain_{DA}}$ are negative, indicating that the current training data fails to yield meaningful gains on these subtasks.

\section{Further Analysis}


\subsection{Impact of Reinforcement Learning}

Since the Group Relative Policy Optimization (GRPO)~\cite{guo2025deepseek} based reinforcement learning (RL) has become a common choice in reasoning training, considerable debate has emerged regarding the respective contributions of Supervised Fine-Tuning (SFT) and RL to reasoning performance~\cite{chu2025sft,yue2025does}. A prominent consensus suggests that RL primarily improves sampling efficiency rather than expanding the fundamental reasoning capability boundary~\cite{yue2025does}. Since Dual Tuning involves only the SFT stage, we further employ an RL training stage based on the Dual-Tuned model with GRPO algorithm~\cite{guo2025deepseek}.

The spatial results (Table~\ref{tab:sp-rl}) show that RL training significantly narrows the gap between DA and CoT modes on spatial subtasks. However, the reasoning efficacy remains unchanged due to the $\mathbf{Gain_{CoT}^{RL}}$ only barely approaches $\mathbf{Gain_{DA}^{RL}}$ on most subtasks. The only exception is the video-based \textit{Route Plan} tasks. Given the training overhead of RL and the higher token cost of CoT inference, DA training is the more suitable choice under the current base model and data conditions.

The mathematical results (Table~\ref{tab:math-rl} in Appendix) show that RL training further magnifies the benefits of CoT finetuning. These improvements do not change the conclusions drawn from Dual Tuning, that reasoning training is currently the more suitable choice. These two contrasting sets of results demonstrate that Dual Tuning provides a practical guide for training decisions.

\subsection{Impact of Thinking Patterns}

The core of CoT data lies in its embedded thinking patterns, which can vary substantially across different sources, whether expert-annotated or model-distilled. Different benchmarks also probe different thinking patterns. For instance, MathVista prioritizes perceptual-logical reasoning with an emphasis on explicit step-by-step trajectories over profound background knowledge. The knowledge-intensive scenarios in MMMU demand reasoning integrated with the model's internal knowledge.

To compare the impact of thinking patterns, we construct a dataset with CoT content distilled from Qwen3-VL-235B~\cite{bai2025qwen3vltechnicalreport}. Figure~\ref{fig:think_pattern} shows that different CoT content leads to markedly different distributions across subtasks. The conciseness of reasoning content, specifically the absence of irrelevant steps or circular reasoning, serves as an important factor in determining the efficacy of a thinking pattern. In Figure~\ref{fig:token}, $\mathbf{Gain_{token}}$ provides an intuitive measure of the efficiency of CoT content. In \textit{Algebraic Reasoning}, the new dataset exhibits higher $\mathbf{Gain_{token}}$ but lower $\mathbf{Gain_{CoT}}$, indicating that the thinking patterns in the new data are more concise.

\subsection{Guidance for Data Curation}

Beyond quantifying the reasoning efficacy of multimodal tasks, Figure~\ref{fig:mmmu_sca} also provides a clear depiction of the gains and losses yielded by both CoT and DA data. A natural question then arises: \textit{Can these results serve as practical guidance for data curation?} 
With subtask labels annotated for each data sample, we conduct a series of experiments to verify that the Dual Tuning results reflect the true contribution of the data. 

In the first set, we partition subtasks into upper and lower halves based on $\mathbf{Gain_{CoT}}$, as shown in Figure~\ref{fig:sca_L}. After training independently on the data belonging to each half, we find that the data corresponding to negative tasks indeed yields negative gains during standalone training, and vice versa. Here we evaluate only the subtasks covered by the corresponding training data labels.

In the second set, we validate the DA data in the same manner.  According to  $\mathbf{Gain_{DA}}$, the left half contains tasks where the corresponding data bring negative gains, and the right half contains tasks with positive gains. After training independently on the data belonging to each half,  Figure~\ref{fig:sca_S} (Appendix) shows that left-side tasks predominantly show negative gains (red markers), while right-side positive tasks mostly achieve positive gains.

In the third set, we conduct two separate Dual Tuning using data from the lower-left negative region and the remaining three positive regions, respectively. Figure~\ref{fig:sca_SL} (Appendix) shows that for lower-left subtasks, training solely on corresponding data predominantly yields negative gains, where the gain is taken as $\mathbf{\max(Gain_{CoT}, Gain_{DA}})$. The subtasks in green and pink positive regions reveal exclusively positive gains.

This evidence confirms that Dual Tuning results capture the true efficacy of training data, offering actionable guidance for the identification and refinement of data subsets.

\section{Conclusion and Limitations}
We propose Dual Tuning as a data curation framework to quantify data and reasoning efficacy for multimodal LLMs training. Our study provides practical guidance for organizing training data and tailoring training strategies across different tasks, base models, and data quality conditions, enabling more resource-efficient Multimodal LLM training. Our study has limitations in covering broader reasoning scenarios such as agentic or GUI tasks due to resource constraints, and we encourage future work to expand the scope.

\clearpage

\normalsize
\bibliography{main}


\newpage
\appendix
\onecolumn
\section{Additional Experimental Results}


\begin{table*}[ht]
  \caption{\textbf{Ming-lite-omni} experimental results on spatial tasks. Values in red and green denote negative and positive results, respectively. A task is identified as suitable for reasoning-oriented training only when both $\mathbf{Gain_{CoT}}$ and $\mathbf{GAP_{DT}}$ exhibit concurrent positive values (highlighted in \textbf{\textcolor{myTextGreenColor}{green}}).}
  \label{tab:ming-sp}
  \setlength{\tabcolsep}{5.0 mm}
  \centering
  \small
  \begin{tabular*}{0.98\linewidth}{l|ccc|ccccc}
  \toprule
  Task & $\mathbf{B_S}$ & $\mathbf{B_L}$ & $\mathbf{GAP_{B}}$ & $\mathbf{DT_S}$ & $\mathbf{DT_L}$ & $\mathbf{Gain_{DA}}$ & \textcolor{myTextGreenColor}{$\mathbf{Gain_{CoT}}$} & \textcolor{myTextGreenColor}{$\mathbf{GAP_{DT}}$} \\
  \midrule
  \textcolor{myTextBenchColor}{VSIBench} & 36.07 & 36.69 & \ratioCell{0.62} & 49.40 & 38.98 & \ratioCell{34.64} & \ratioCell{6.24} & \ratioCell{-10.42} \\
  \midrule
  \cellcolor{orange!10} Obj. Count & 40.23 & 40.32 & \ratioCell{0.09} & 51.98 & 29.04 & \ratioCell{28.92} & \ratioCell{-27.98} & \ratioCell{-22.94} \\
  \cellcolor{orange!10} Abs. Dist. & 13.68 & 17.22 & \ratioCell{3.54} & 37.53 & 25.22 & \ratioCell{117.94} & \ratioCell{46.46} & \ratioCell{-12.31} \\
  \cellcolor{orange!10} Obj. Size & 43.89 & 45.62 & \ratioCell{1.73} & 66.14 & 54.39 & \ratioCell{44.98} & \ratioCell{19.22} & \ratioCell{-11.75} \\
  \cellcolor{orange!10} Room Size & 31.77 & 29.06 & \ratioCell{-2.71} & 53.16 & 29.93 & \ratioCell{67.33} & \ratioCell{-5.79} & \ratioCell{-23.23} \\
  \cellcolor{orange!10} Rel. Dist. & 45.07 & 44.79 & \ratioCell{-0.28} & 52.54 & 42.82 & \ratioCell{16.57} & \ratioCell{-4.99} & \ratioCell{-9.72} \\
  \cellcolor{orange!10} Rel. Dir. & 43.63 & 45.61 & \ratioCell{1.98} & 41.12 & 45.21 & \ratioCell{-9.84} & \ratioCell{-0.88} & \ratioCell{4.09} \\
  \cellcolor{orange!10} Route Plan & 29.38 & 33.51 & \ratioCell{4.13} & 30.93 & 31.96 & \ratioCell{-7.70} & \ratioCell{-4.63} & \ratioCell{1.03} \\
  \cellcolor{orange!10} Appr. Order & 40.94 & 37.38 & \ratioCell{-3.56} & 61.81 & 53.24 & \ratioCell{50.98} & \ratioCell{30.04} & \ratioCell{-8.57} \\
  \midrule
  \textcolor{myTextBenchColor}{CVBench} & 74.53 & 76.69 & \ratioCell{2.16} & 84.99 & 81.05 & \ratioCell{10.82} & \ratioCell{5.69} & \ratioCell{-3.94} \\
  \midrule
  \cellcolor{cyan!10} Count & 68.91 & 65.23 & \ratioCell{-3.68} & 72.21 & 66.50 & \ratioCell{4.79} & \ratioCell{-3.50} & \ratioCell{-5.71} \\
  \cellcolor{cyan!10} Relation & 81.23 & 89.08 & \ratioCell{7.85} & 95.38 & 90.15 & \ratioCell{7.07} & \ratioCell{1.20} & \ratioCell{-5.23} \\
  \cellcolor{cyan!10} Depth & 84.17 & 78.33 & \ratioCell{-5.84} & 90.67 & 87.67 & \ratioCell{7.72} & \ratioCell{4.16} & \ratioCell{-3.00} \\
  \cellcolor{cyan!10} Distance & 65.00 & 76.67 & \ratioCell{11.67} & 84.83 & 83.67 & \ratioCell{10.64} & \ratioCell{9.13} & \ratioCell{-1.16} \\
  \bottomrule
\end{tabular*}
\end{table*}


\begin{table*}[ht]
  \caption{\textbf{Ming-lite-omni} experimental results on MathVista tasks. Values in red and green denote negative and positive results, respectively. A task is identified as suitable for reasoning-oriented training only when both $\mathbf{Gain_{CoT}}$ and $\mathbf{GAP_{DT}}$ exhibit concurrent positive values (highlighted in \textbf{\textcolor{myTextGreenColor}{green}}).}
  \label{tab:ming-mathvista}
  \small
  \setlength{\tabcolsep}{5.1 mm}
  \centering
  \begin{tabular*}{0.98\linewidth}{l|ccc|ccccc}
  \toprule
  Task & $\mathbf{B_S}$ & $\mathbf{B_L}$ & $\mathbf{GAP_{B}}$ & $\mathbf{DT_S}$ & $\mathbf{DT_L}$ & $\mathbf{Gain_{DA}}$ & \textcolor{myTextGreenColor}{$\mathbf{Gain_{CoT}}$} & \textcolor{myTextGreenColor}{$\mathbf{GAP_{DT}}$} \\
  \midrule
  \textcolor{myTextBenchColor}{MathVista} & 72.17 & 79.00 & \ratioCell{6.83} & 70.80 & 78.77 & \ratioCell{-10.38} & \ratioCell{-0.29} & \ratioCell{7.97} \\
  \cellcolor{yellow!10} Comm. & 52.31 & 49.77 & \ratioCell{-2.54} & 50.46 & 48.61 & \ratioCell{-3.54} & \ratioCell{-7.07} & \ratioCell{-1.85} \\
  \cellcolor{yellow!10} Scientific & 72.62 & 77.68 & \ratioCell{5.06} & 75.89 & 75.60 & \ratioCell{-2.30} & \ratioCell{-2.68} & \ratioCell{-0.29} \\
  \cellcolor{yellow!10} Geometry & 82.33 & 87.79 & \ratioCell{5.46} & 79.60 & 88.07 & \ratioCell{-9.33} & \ratioCell{0.32} & \ratioCell{8.47} \\
  \cellcolor{yellow!10} Arithmetic & 73.48 & 89.85 & \ratioCell{16.37} & 69.55 & 89.70 & \ratioCell{-22.59} & \ratioCell{-0.17} & \ratioCell{20.15} \\
  \cellcolor{yellow!10} Logical & 12.61 & 27.03 & \ratioCell{14.42} & 15.32 & 29.73 & \ratioCell{-43.32} & \ratioCell{9.99} & \ratioCell{14.41} \\
  \cellcolor{yellow!10} Statistical & 84.47 & 88.31 & \ratioCell{3.84} & 82.90 & 89.70 & \ratioCell{-6.13} & \ratioCell{1.57} & \ratioCell{6.80} \\
  \cellcolor{yellow!10} Algebraic & 72.40 & 80.21 & \ratioCell{7.81} & 76.04 & 76.56 & \ratioCell{-5.20} & \ratioCell{-4.55} & \ratioCell{0.52} \\

  \bottomrule
\end{tabular*}
\end{table*}

\begin{table*}[ht]
  \caption{Performance Comparison Following Subsequent RL Training on Dual-Tuned Models for MathVista Tasks. Values in red and green denote negative and positive results, respectively. A task is identified as suitable for reasoning-oriented training only when both $\mathbf{Gain_{CoT}}$ and $\mathbf{GAP_{DT}}$ exhibit concurrent positive values (highlighted in \textbf{\textcolor{myTextGreenColor}{green}}).}
  \label{tab:math-rl}
  \setlength{\tabcolsep}{7.1 mm}
  \centering
  \small
  \begin{tabular*}{0.98\linewidth}{l|ccc|ccccc}
  \toprule
  Task & $\mathbf{Gain_{DA}}$ & \textcolor{myTextGreenColor}{$\mathbf{Gain_{CoT}}$} & \textcolor{myTextGreenColor}{$\mathbf{GAP_{DT}}$} & $\mathbf{Gain_{DA}^{RL}}$ & \textcolor{myTextGreenColor}{$\mathbf{Gain_{CoT}^{RL}}$} & \textcolor{myTextGreenColor}{$\mathbf{GAP_{DT}^{RL}}$} \\
  \midrule
  \textcolor{myTextBenchColor}{MathVista} & \ratioCell{-3.52} & \ratioCell{4.35} & \ratioCell{5.50} & \ratioCell{-3.58} & \ratioCell{6.07} & \ratioCell{6.74} \\
  \midrule
  \cellcolor{yellow!10} Comm. & \ratioCell{10.42} & \ratioCell{3.65} & \ratioCell{-3.01} & \ratioCell{11.48} & \ratioCell{11.99} & \ratioCell{0.23} \\
  \cellcolor{yellow!10} Scientific & \ratioCell{0.00} & \ratioCell{0.00} & \ratioCell{0.00} & \ratioCell{0.40} & \ratioCell{0.80} & \ratioCell{0.30} \\
  \cellcolor{yellow!10} Geometry & \ratioCell{-9.17} & \ratioCell{5.70} & \ratioCell{10.49} & \ratioCell{-10.19} & \ratioCell{7.53} & \ratioCell{12.50} \\
  \cellcolor{yellow!10} Arithmetic & \ratioCell{-11.1} & \ratioCell{2.08} & \ratioCell{10.60} & \ratioCell{-11.49} & \ratioCell{2.26} & \ratioCell{11.06} \\
  \cellcolor{yellow!10} Logical & \ratioCell{-34.37} & \ratioCell{-18.77} & \ratioCell{4.50} & \ratioCell{-34.37} & \ratioCell{0.00} & \ratioCell{9.91} \\
  \cellcolor{yellow!10} Statistical & \ratioCell{1.44} & \ratioCell{3.30} & \ratioCell{1.57} & \ratioCell{0.63} & \ratioCell{3.92} & \ratioCell{2.79} \\
  \cellcolor{yellow!10} Algebraic & \ratioCell{-6.61} & \ratioCell{8.82} & \ratioCell{10.93} & \ratioCell{-1.47} & \ratioCell{8.09} & \ratioCell{6.77} \\
  \bottomrule
\end{tabular*}
\end{table*}

\clearpage

\begin{table*}[ht]
  \caption{\textbf{Ming-lite-omni} experimental results on MMMU tasks. Values in red and green denote negative and positive results, respectively. A task is identified as suitable for reasoning-oriented training only when both $\mathbf{Gain_{CoT}}$ and $\mathbf{GAP_{DT}}$ exhibit concurrent positive values (highlighted in \textbf{\textcolor{myTextGreenColor}{green}}).}
  \label{tab:ming-mmmu}
  \setlength{\tabcolsep}{4.5 mm}
  \centering
  \small
  \begin{tabular*}{0.98\linewidth}{l|ccc|ccccc}
  \toprule
  Task & $\mathbf{B_S}$ & $\mathbf{B_L}$ & $\mathbf{GAP_{B}}$ & $\mathbf{DT_S}$ & $\mathbf{DT_L}$ & $\mathbf{Gain_{DA}}$ & \textcolor{myTextGreenColor}{$\mathbf{Gain_{CoT}}$} & \textcolor{myTextGreenColor}{$\mathbf{GAP_{DT}}$} \\
  \midrule
  \textcolor{myTextBenchColor}{MMMU} & 52.89 & 56.44 & \ratioCell{3.55} & 53.11 & 59.11 & \ratioCell{-5.9} & \ratioCell{4.73} & \ratioCell{6} \\
  \midrule
  \cellcolor{blue!10} Art & 66.67 & 70.00 & \ratioCell{3.33} & 66.67 & 66.67 & \ratioCell{-4.76} & \ratioCell{-4.76} & \ratioCell{0.00} \\
  \cellcolor{blue!10} Design & 83.33 & 76.67 & \ratioCell{-6.66} & 70.00 & 70.00 & \ratioCell{-16.00} & \ratioCell{-16.00} & \ratioCell{0.00} \\
  \cellcolor{blue!10} Music & 40.00 & 26.67 & \ratioCell{-13.33} & 46.67 & 16.67 & \ratioCell{16.68} & \ratioCell{-58.33} & \ratioCell{-30.00} \\
  \cellcolor{blue!10} Art Theory & 83.33 & 70.00 & \ratioCell{-13.33} & 80.00 & 80.00 & \ratioCell{-4.00} & \ratioCell{-4.00} & \ratioCell{0.00} \\
  \cellcolor{red!10} Accounting & 46.67 & 66.67 & \ratioCell{20.00} & 46.67 & 73.33 & \ratioCell{-30.00} & \ratioCell{9.99} & \ratioCell{26.66} \\
  \cellcolor{red!10} Finance & 33.33 & 56.67 & \ratioCell{23.34} & 36.67 & 63.33 & \ratioCell{-35.29} & \ratioCell{11.75} & \ratioCell{26.66} \\
  \cellcolor{red!10} Economics & 46.67 & 70.00 & \ratioCell{23.33} & 50.00 & 73.33 & \ratioCell{-28.57} & \ratioCell{4.76} & \ratioCell{23.33} \\
  \cellcolor{red!10} Manage & 40.00 & 60.00 & \ratioCell{20.00} & 46.67 & 50.00 & \ratioCell{-22.22} & \ratioCell{-16.67} & \ratioCell{3.33} \\
  \cellcolor{red!10} Marketing & 53.33 & 80.00 & \ratioCell{26.67} & 63.33 & 86.67 & \ratioCell{-20.84} & \ratioCell{8.34} & \ratioCell{23.34} \\
  \cellcolor{yellow!10} Math & 50.00 & 56.67 & \ratioCell{6.67} & 26.67 & 50.00 & \ratioCell{-52.94} & \ratioCell{-11.77} & \ratioCell{23.33} \\
  \cellcolor{yellow!10} Biology & 40.00 & 40.00 & \ratioCell{0.00} & 46.67 & 46.67 & \ratioCell{16.68} & \ratioCell{16.68} & \ratioCell{0.00} \\
  \cellcolor{yellow!10} Chemistry & 30.00 & 43.33 & \ratioCell{13.33} & 36.67 & 53.33 & \ratioCell{-15.37} & \ratioCell{23.08} & \ratioCell{16.66} \\
  \cellcolor{yellow!10} Geography & 63.33 & 43.33 & \ratioCell{-20.00} & 56.67 & 50.00 & \ratioCell{-10.52} & \ratioCell{-21.05} & \ratioCell{-6.67} \\
  \cellcolor{yellow!10} Physics & 50.00 & 60.00 & \ratioCell{10.00} & 43.33 & 60.00 & \ratioCell{-27.78} & \ratioCell{0.00} & \ratioCell{16.67} \\
  \cellcolor{cyan!10} Basic Med. Sci. & 56.67 & 66.67 & \ratioCell{10.00} & 63.33 & 70.00 & \ratioCell{-5.01} & \ratioCell{4.99} & \ratioCell{6.67} \\
  \cellcolor{cyan!10} Clinical Med. & 63.33 & 56.67 & \ratioCell{-6.66} & 60.00 & 63.33 & \ratioCell{-5.26} & \ratioCell{0.00} & \ratioCell{3.33} \\
  \cellcolor{cyan!10} Diag. \& Lab. Med. & 43.33 & 33.33 & \ratioCell{-10.00} & 30.00 & 40.00 & \ratioCell{-30.76} & \ratioCell{-7.69} & \ratioCell{10.00} \\
  \cellcolor{cyan!10} Public Health & 66.67 & 83.33 & \ratioCell{16.66} & 63.33 & 83.33 & \ratioCell{-24.00} & \ratioCell{0.00} & \ratioCell{20.00} \\
  \cellcolor{cyan!10} Pharmacy & 66.67 & 80.00 & \ratioCell{13.33} & 63.33 & 73.33 & \ratioCell{-20.84} & \ratioCell{-8.34} & \ratioCell{10.00} \\
  \cellcolor{green!10} History & 63.33 & 66.67 & \ratioCell{3.34} & 70.00 & 83.33 & \ratioCell{4.99} & \ratioCell{24.99} & \ratioCell{13.33} \\
  \cellcolor{green!10} Literature & 80.00 & 90.00 & \ratioCell{10.00} & 80.00 & 83.33 & \ratioCell{-11.11} & \ratioCell{-7.41} & \ratioCell{3.33} \\
  \cellcolor{green!10} Psychology & 63.33 & 56.67 & \ratioCell{-6.66} & 56.67 & 56.67 & \ratioCell{-10.52} & \ratioCell{-10.52} & \ratioCell{0.00} \\
  \cellcolor{green!10} Sociology & 66.67 & 53.33 & \ratioCell{-13.34} & 66.67 & 60.00 & \ratioCell{0.00} & \ratioCell{-10.00} & \ratioCell{-6.67} \\
  \cellcolor{orange!10} Agriculture & 40.00 & 53.33 & \ratioCell{13.33} & 53.33 & 46.67 & \ratioCell{0.00} & \ratioCell{-12.49} & \ratioCell{-6.66} \\
  \cellcolor{orange!10} Arch. \& Eng. & 36.67 & 26.67 & \ratioCell{-10.00} & 36.67 & 33.33 & \ratioCell{0.00} & \ratioCell{-9.11} & \ratioCell{-3.34} \\
  \cellcolor{orange!10} Comp. Sci. & 53.33 & 56.67 & \ratioCell{3.34} & 63.33 & 60.00 & \ratioCell{11.75} & \ratioCell{5.88} & \ratioCell{-3.33} \\
  \cellcolor{orange!10} Electronics & 40.00 & 33.33 & \ratioCell{-6.67} & 43.33 & 43.33 & \ratioCell{8.33} & \ratioCell{8.33} & \ratioCell{0.00} \\
  \cellcolor{orange!10} Energy \& Power & 46.67 & 40.00 & \ratioCell{-6.67} & 60.00 & 53.33 & \ratioCell{28.56} & \ratioCell{14.27} & \ratioCell{-6.67} \\
  \cellcolor{orange!10} Materials & 36.67 & 40.00 & \ratioCell{3.33} & 33.33 & 40.00 & \ratioCell{-16.68} & \ratioCell{0.00} & \ratioCell{6.67} \\
  \cellcolor{orange!10} Mech. Eng. & 36.67 & 36.67 & \ratioCell{0.00} & 33.33 & 43.33 & \ratioCell{-9.11} & \ratioCell{18.16} & \ratioCell{10.00} \\

  \bottomrule
\end{tabular*}
\end{table*}

\clearpage

\begin{figure*}[!ht]
  \vspace{-0.3cm}
  \centering
  \includegraphics[width=0.98\linewidth]{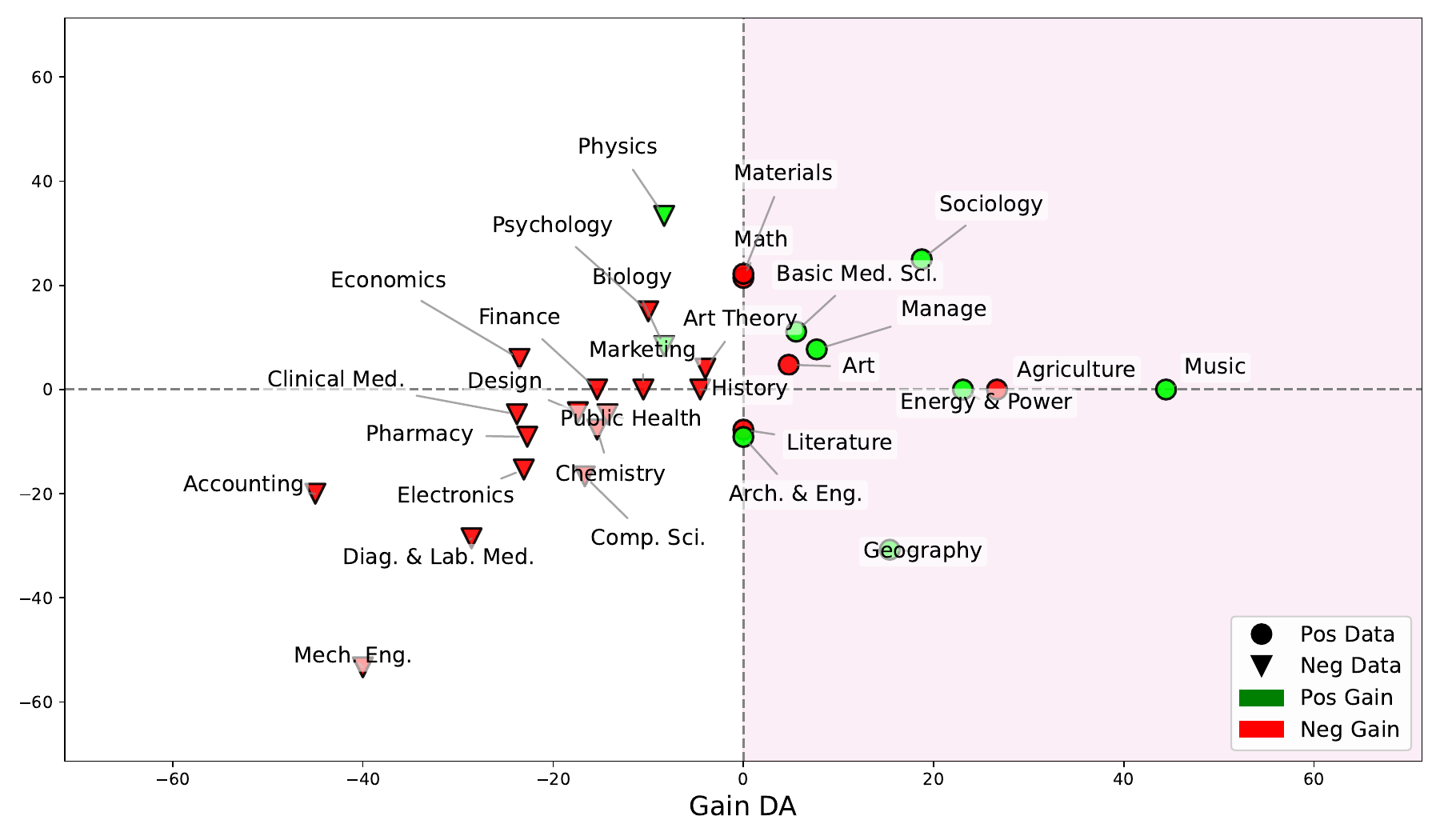}
  \vspace{-0.4cm}
  \caption{We partition tasks into two halves using $\mathbf{Gain_{DA}}$ from Figure~\ref{fig:mmmu_sca} and conduct two separate DA training on the data belonging to each half. The results show that left-side tasks predominantly show negative gains and right-side positive tasks mostly achieve positive gains after standalone training, which confirms the efficacy of the corresponding data.}
  \label{fig:sca_S}
\end{figure*}

\begin{figure*}[!ht]
  \centering
  \includegraphics[width=0.98\linewidth]{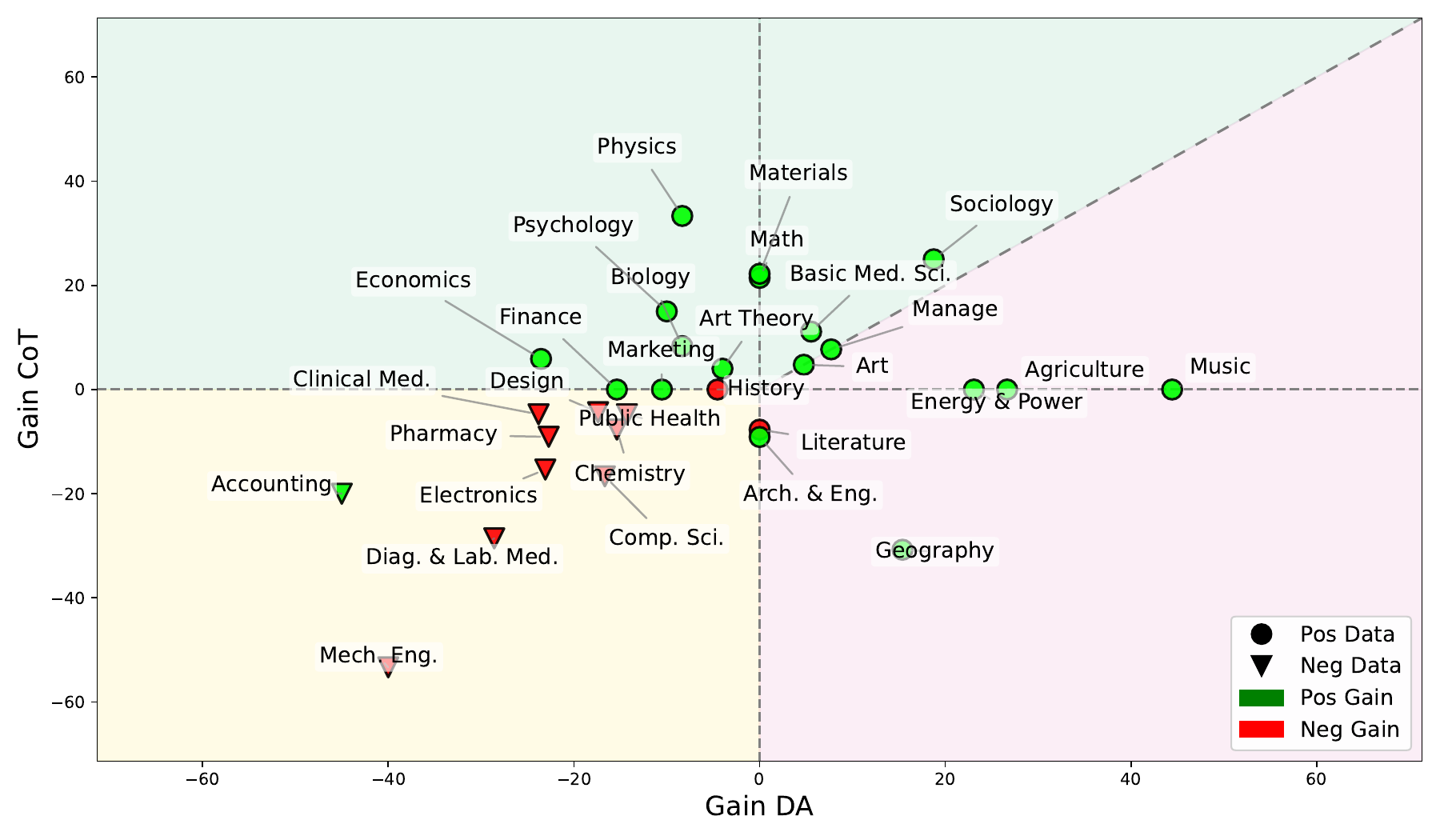}
  \vspace{-0.4cm}
  \caption{We separately train models with data from the lower-left negative region and the remaining three positive regions. For tasks in the lower-left yellow region, training solely on corresponding data predominantly yields negative gains. For the green and pink positive regions, training on the corresponding data reveals exclusively positive gains.}
  \label{fig:sca_SL}
\end{figure*}


\end{document}